\documentclass[11pt]{article}

\usepackage[preprint]{acl}

\usepackage{times}
\usepackage{latexsym}

\usepackage[T1]{fontenc}
\usepackage[utf8]{inputenc}

\usepackage{microtype}

\usepackage{inconsolata}

\usepackage{graphicx}
\usepackage{enumitem}
\usepackage{booktabs}

\usepackage{amsmath}
\usepackage{amssymb}
\usepackage{algorithm}
\usepackage{algpseudocode}
\usepackage{booktabs}

\title{AgentKGV: Agentic LLM-RAG Framework with Two-Stage Training for the Fact Verification of Knowledge Graphs}

\author{
 \textbf{Yumin Heo\textsuperscript{1}},
 \textbf{Hyeon-gu Lee\textsuperscript{2}},
 \textbf{Sumin Seo\textsuperscript{2}},
 \textbf{Youngjoong Ko\textsuperscript{1}}\thanks{Corresponding author}
\\
 \textsuperscript{1}SungKyunKwan University,
 \textsuperscript{2}NAVER,
\\
 \texttt{ymheo1123@gmail.com, yjko@skku.edu}
\\
 \texttt{\{hyeongu.lee, sumin.seo\}@navercorp.com}
}

\begin{document}
\maketitle
\begin{abstract}
Knowledge graphs (KGs) are often automatically constructed from large-scale corpora, but they inevitably contain factual errors due to noisy sources and extraction failures, and verifying them reliably at industrial scale remains a critical challenge. To address this, we propose AgentKGV, the Agentic LLM-RAG framework for KG fact Verification, that integrates dynamic routing and iterative query rewriting, which handles surface-form mismatch in document-level retrieval. To make this framework more accurate and cost-efficient for industrial deployment, we further introduce a two-stage training strategy: turn-level distillation-based SFT that transfers reasoning ability from a large teacher model into a small model for stable query rewriting and reasoning, and trajectory-level GRPO that optimizes the search policy to reduce unnecessary retrieval at scale. On the long-tail-predicate split of the open-domain T-REx benchmark, our framework improves macro-F1 over single-turn RAG by 5.5 \%p, and two-stage training does it further by 9.4 \%p. GRPO also cuts the average number of search calls from 3.24 to 1.63 without lowering accuracy.
\end{abstract}

\section{Introduction}
\label{sec:intro}
Knowledge graphs encode entities and their relationships as triples in the form (\textit{subject}, \textit{predicate}, \textit{object}), and they serve as core components across a wide range of knowledge-intensive applications such as search engines, recommendation systems, question answering, and decision support. In industrial settings, the demand for automatic construction of KGs from large volumes of documents has considerably grown. However, automatically constructed KGs inevitably contain incorrect information due to ambiguous sentence structures, low-reliability sources, and errors in NLP models. Therefore, the factual validity of extracted triples has become a critical bottleneck for the downstream reliability of industrial KGs.

Researchers have traditionally relied on isolated methodologies to evaluate the validity of automatically constructed triples. Graph-based methods \cite{transE} assess validity through structural consistency, but they often fail to detect real-world factual errors because they rely solely on internal graph topology. LLM-based methods \cite{Pan_2024} offer broader semantic reasoning, but they remain vulnerable to domain-specific hallucinations. RAG-based methods \cite{lewis2020retrieval, trivedi2022interleaving} utilize external document retrieval, but they fail when the system cannot retrieve the relevant information. A further difficulty is the structural modality gap between compressed triples and natural language documents, because factual information in documents rarely appears in the same standardized form as KG triples and this makes single-round retrieval unreliable.

To overcome these limitations, recent research has adopted Agentic LLM frameworks \cite{yao2022react, schick2023toolformer}, in which autonomous agents dynamically orchestrate reasoning and retrieval. On this paradigm, we propose AgentKGV in which the agent first decides whether it verifies a triple through internal parametric knowledge or through external retrieval through dynamic routing. When external retrieval is necessary, the agent then iteratively rewrites an input triple into natural language queries that are compatible with document-level retrieval. Because this iterative rewriting transforms the compressed triple into diverse natural language expressions, it grounds the verification in retrieved evidence.

However, deploying this framework reliably and efficiently in industrial KGs requires addressing two practical challenges. First, query rewriting can become unstable because the model may lack a stable semantic anchor for domain-specific predicates, and industrial KGs contain a large proportion of long-tail predicates that are rarely observed in general pretraining corpora. Second, the iterative search policy can become inefficient because the model may not know when to stop searching, and this inefficiency is especially costly at industrial scale where even a small increase in search iterations per triple leads to substantial computational overhead.

To make the framework more accurate and cost-efficient for industrial deployment, we introduce a two-stage training strategy. The first stage applies distillation-based SFT, in which a large teacher model generates complete verification trajectories. From each successful trajectory, we take the final judgment together with the query that immediately precedes it, and we train the small model on these. The query-rewriting part teaches the model how the teacher rewrites a triple into an effective query, and the judgment part teaches the model how the teacher reasons over the retrieved evidence to reach a conclusion. Through this distillation, the small model learns the query rewriting and reasoning ability of the teacher, and it acquires a stable semantic foundation for triple-grounded query rewriting and routing, even on domain-specific predicates. The second stage applies trajectory-level GRPO, in which the model is trained through full rollouts with trajectory-level rewards. We additionally impose a per-turn search penalty, so that the model is discouraged from issuing unnecessary search calls. Through this design, the model learns when to stop searching and how to draw a conclusion under insufficient evidence. The contributions of this work are summarized as follows:

\begin{itemize}

\item We propose \textbf{AgentKGV}, the Agentic LLM-RAG framework for KG fact verification that integrates dynamic routing and iterative query rewriting to verify triples against document-level evidence through multi-turn retrieval.

\item We introduce a \textbf{two-stage training strategy} that combines turn-level distillation-based SFT and trajectory-level GRPO, and this makes the framework more accurate and cost-efficient for industrial deployment.

\item We validate AgentKGV on an open-domain English benchmark, confirming its effectiveness on \textbf{long-tail predicates} rarely seen during training, and we further verify its applicability on a \textbf{real-world Korean enterprise KG}.

\end{itemize}

\begin{figure*}[t]
    \centering
    \includegraphics[width=\textwidth]{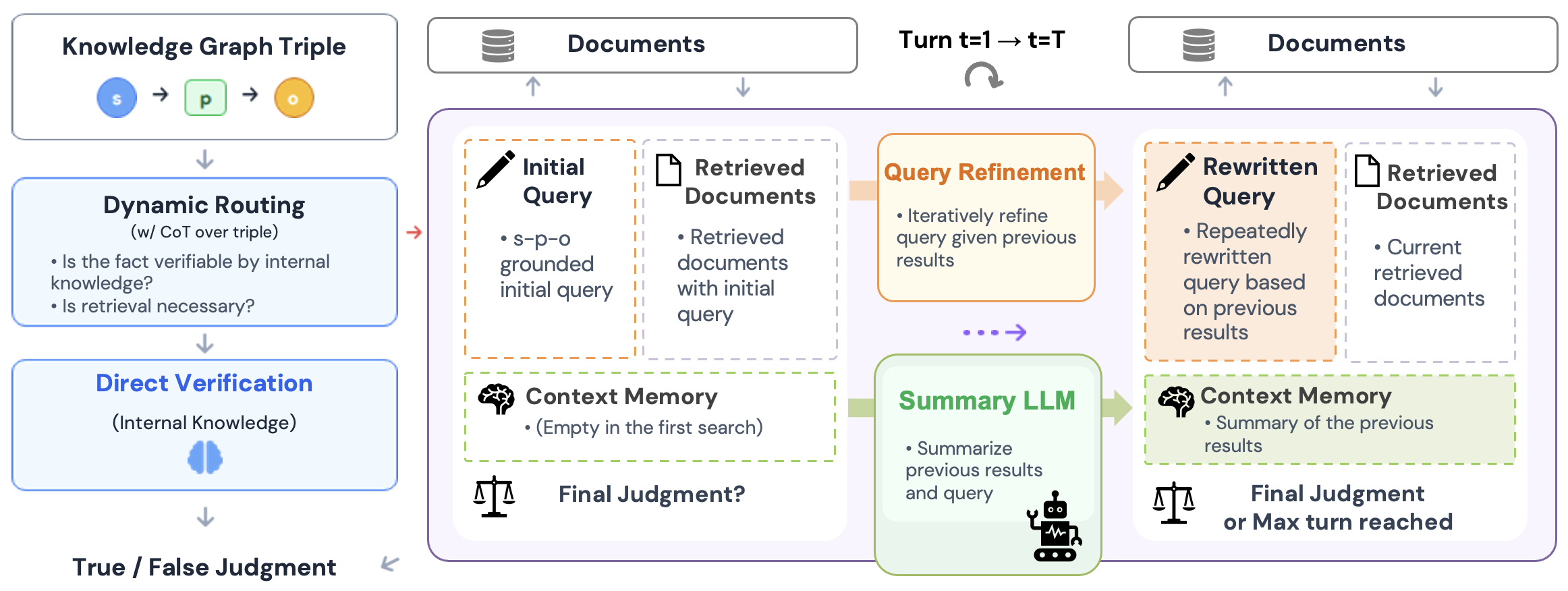}
    \caption{Overview of the proposed Agentic LLM-RAG framework. The agent dynamically routes between internal verification and external retrieval, and iteratively rewrites the triple into retrieval-compatible queries until it produces a final judgment.}
    \label{fig1:top}
\end{figure*}

\section{Related Work}
\subsection{KG Fact Verification}
Approaches to verifying KG triple validity fall into three groups. \textit{Graph-based methods} assess validity from internal graph structure, either by scoring triples with learned embeddings \cite{transE, yang2015embedding} or by path- and reasoning-based inference over the graph \cite{lao2011random, das2017go}. They capture structural regularities but rely solely on graph topology, and thus cannot detect construction-time factual errors that have no support within the graph. \textit{LLM-based methods} \cite{Pan_2024} instead exploit the parametric knowledge of large language models, and they offer broader semantic coverage but remain vulnerable to hallucination on domain-specific predicates that are rarely seen during pretraining. \textit{RAG-based methods} ground the judgment in retrieved documents \cite{lewis2020retrieval, popat2018declare, chen2022generating,pan2023fact}, and recent work extends them with iterative \cite{trivedi2022interleaving, asai2023self} and adaptive \cite{mallen2022not, shi2024generate} retrieval. These reduce hallucination but are designed for natural-language question answering, and they overlook a challenge specific to KG verification: the target is a compressed $(s, p, o)$ triple rather than a question, and a single retrieval over it is often unreliable because documents express the same fact with different wording.

\textit{Agentic LLM frameworks} go one step further. The model acts as an autonomous agent that interleaves reasoning with tool or retrieval calls \cite{yao2022react, schick2023toolformer}, and recent work trains the agent with reinforcement learning so that it learns when and how to search \cite{jin2025search, song2025r1}. These frameworks are effective on multi-hop question answering, but they are not designed for KG triple verification, and they do not address the $(s, p, o)$-to-document modality gap. Our framework adopts the agentic paradigm for KG fact verification, and it addresses this gap through dynamic routing and $(s, p, o)$-grounded iterative query rewriting.

\section{Methodology}

\subsection{Problem Formulation}

We formulate KG fact verification as a binary classification task. Given a triple $\tau = (s, p, o)$, the goal is to assign a label $y \in \{\text{true}, \text{false}\}$, which indicates whether the triple is factually correct, to the triple. The verification is performed over a document corpus $\mathcal{D} = \{d_1, d_2, \ldots, d_N\}$, where each document may contain evidences relevant to the triple.

\subsection{Agentic LLM-RAG Framework}
\subsubsection{Overall Architecture}
The proposed framework operates as a multi-turn agent that interacts with a retrieval system. Figure~\ref{fig1:top} shows the overview of the framework. Given a triple $\tau$, the agent first performs dynamic routing and decides whether to verify the triple directly through its internal knowledge or to initiate iterative retrieval. When retrieval is initiated, the agent selects one of several component combinations as a guide and writes an initial query, and then it enters a multi-turn loop.

At each turn $t$, the agent observes a context $c_t$ that consists of the triple, a running summary of previous turns, the most recent query, and its retrieval result: \begin{equation} c_t = \big[\tau,\ \mathcal{S}_{<t},\ q_{t-1},\ \mathcal{R}_{t-1}\big] \end{equation} where $\mathcal{S}_{<t}$ is the running summary of earlier turns, and $q_{t-1}$ and $\mathcal{R}_{t-1}$ are the most recent query and its retrieved documents. Based on this context, the agent decides whether the accumulated evidence is sufficient to produce a final judgment or whether it should rewrite the query and continue searching. The loop continues until the agent produces a final judgment, or until it reaches the maximum number of turns $T$. Across these turns, a separate summarization module compresses the retrieval history so that the context remains compact.

\subsubsection{Dynamic Routing}
The routing decision is made through explicit chain-of-thought reasoning over a triple. When the triple can be confidently verified through internal parametric knowledge, the agent produces a judgment without retrieval. Otherwise, the agent initiates iterative retrieval instead. This adaptive routing avoids unnecessary retrieval and reduces cost.

\subsubsection{$(s, p, o)$-grounded Initial Query Writing}
When retrieval is necessary, the agent transforms the compressed triple into a natural language query that is compatible with document-level retrieval. Unlike general query rewriting that operates on a natural language query, the agent can leverage the explicit $(s, p, o)$ structure of the triple. Actually, document-level retrieval using only a single fixed query can be unreliable; an overly broad query returns documents that mention the entity but do not include the fact, while an overly specific query may match no document when the corpus expresses the same fact with different wording. In our method, a set of component combinations (Table~\ref{tab:query_combinations} in Appendix~\ref{sec:appendix-query-combinations}) is presented to the agent as selectable options in the prompt, and one combination is chosen per each turn to set the specificity and surface form of its query. This enable the agent to adjust retrieval granularity and cover lexical variation of the predicate.

\subsubsection{Iterative Query Rewriting}
The agent refines the query multiple times until evidence is found. Starting from the second turn, the agent no longer writes each query independently. Instead, this system refines subsequent queries based on previous queries and their results to enable it to find relevant evidence more accurately. When the previous result is only partially relevant, the agent reformulates the query toward a more suitable component combination or surface form. On the other hand, when the previous query retrieves no useful documents, the agent changes the formulation rather than repeats the same search. This feedback between retrieval results and subsequent queries enables the agent to more effectively search for evidences that support or refute the triple. The iteration continues until the agent retrieves sufficient evidences for a final judgment, or until it reaches the maximum number of turns.

\subsubsection{Retrieval Summarization Module}
The retrieval summarization module maintains the running summary $\mathcal{S}_{<t}$ across turns. At each turn, it combines the previous summary with the most recent query and retrieval result: \begin{equation} \mathcal{S}_{<t+1} = \text{Summarizer}\big(\mathcal{S}_{<t},\ q_{t},\ \mathcal{R}_{t}\big) \end{equation} In this way, the full result of the current turn is compressed into the summary at the next turn, so that the agent always sees the compressed history of earlier turns together with the full result of only the current turn.
This design provides two benefits. First, it bounds the context length, because all but the most recent turn are stored in compressed form rather than as full documents. Second, the summary makes the search history explicit, so the agent can avoid redundant queries and instead target the evidence that is still missing. Therefore, the module maintains the context compact while it supports more effective query rewriting across turns.

\begin{figure*}[t]
    \centering
    \includegraphics[width=\textwidth]{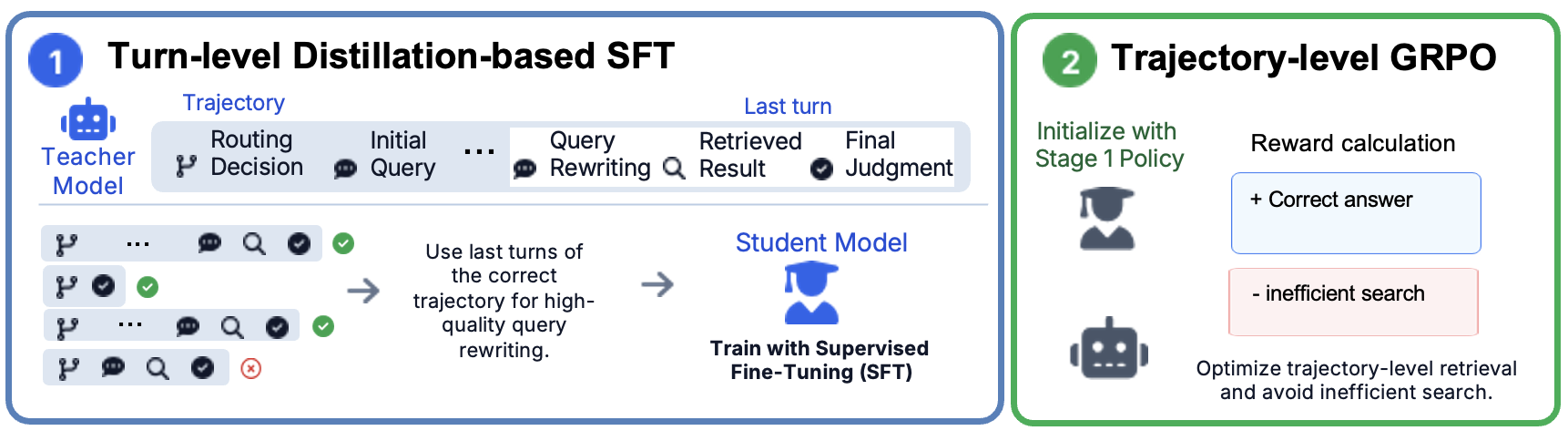}
    \caption{Overview of the two-stage training strategy.
    \textbf{Stage~1 (Distillation-based SFT)} distills the last turn of correct teacher trajectories into two state--action pairs that supervise query rewriting and evidence-grounded reasoning. \textbf{Stage~2 (GRPO)} optimizes the search policy with a trajectory-level reward combining final correctness and a per-turn search penalty, reducing redundant retrieval while preserving accuracy.}
    \label{fig:training}
\end{figure*}

\subsection{Two-Stage Training Strategy}
To deploy the framework reliably and efficiently in industrial KGs, a two-stage sequential strategy is presented for the two practical challenges in Section~\ref{sec:intro}: unstable query rewriting on domain-specific predicates and an inefficient search policy. 

\subsubsection{Stage 1: Distillation-based SFT}

The first stage stabilizes query rewriting and routing through supervised fine-tuning. Because a small target model has limited reasoning capacity, we use a large teacher model to generate complete verification trajectories. To bridge the knowledge gap between the two models, we configure the teacher model to invoke the retrieval function whenever the triple is uncertain, and to provide direct answers only for triples that can be clearly verified using general knowledge. This prevents the target model from relying too heavily on its limited parametric knowledge.

We retain only the trajectories whose final judgment matches the gold label, and then we take the last turn of these trajectories. This turn decomposes into two parts: query rewriting and final judgment. The query-rewriting part retrieves the document using the rewritten query; we treat it as a successful rewritten query, since no further search follows. The judgment part produces the correct label, which serves as supervision for reasoning over the retrieved evidence. From these, we form two state–action pairs, $(s_{t-1}, a_{t-1})$ and $(s_{t-2}, a_{t-2})$, where $s_{t-1}$ and $s_{t-2}$ follows the inference-time context format of the query rewriting part and the final judgement part, respectively; the first pair supervises query rewriting and the second one does answer generation. The training objective is the standard SFT loss: \begin{equation} \mathcal{L}_{\text{SFT}} = -\mathbb{E}_{(s_t, a_t) \sim \mathcal{D}_{\text{train}}}\big[\log P_\theta(a_t \mid s_t)\big] \end{equation}
The retrieved documents and summaries appear only in $s_t$ and are excluded from the loss, because the retrieval and summarization modules are not trained in our model.

\subsubsection{Stage 2: GRPO for Trajectory-level Search Policy Optimization}
The second stage optimizes the iterative search policy through trajectory-level reinforcement learning. Turn-level SFT stabilizes individual decisions but cannot capture the global trade-off between search depth and verification accuracy; continuing to search is always locally reasonable when evidence is insufficient, but excessive iterations can harm both efficiency and accuracy. Therefore, we apply Group Relative Policy Optimization (GRPO) \cite{shao2024deepseekmath, deepseek2025} with trajectory-level rewards.

For each triple, the model generates a group of full trajectories that follow the same context structure as inference. The trajectory-level reward combines final correctness and search cost terms: \begin{equation}
R = R_{\text{correct}} + R_{\text{search}}
\end{equation}
The correctness term assigns $+1.0$ when the final judgment matches the gold label, $0$ when it does not, and $-0.5$ when the output cannot be parsed into a valid label.

The search cost term penalizes unnecessary retrieval. The first search incurs no penalty, because at least one retrieval is generally required to ground the
verification, and each subsequent search incurs a fixed penalty:
\begin{equation}
R_{\text{search}} = -\alpha \cdot \max(0,\ N_{\text{search}} - 1)
\end{equation}
where $\alpha$ is the small positive number, and $N_{\text{search}}$ is the total number of retrieval calls in the
trajectory.

Within each group, the advantage of the $i$-th trajectory $A_i$ is computed
by standardizing $R_i$ against the group's mean and standard deviation
\cite{shao2024deepseekmath}. This trajectory-level advantage is broadcasted
to every model-generated token, and the policy is updated by maximizing the
standard clipped GRPO objective with a KL penalty against the reference model.
Tokens produced by the retrieval environment (i.e., the returned documents)
are excluded from the gradient, so the policy gradient is computed only over
the model's own reasoning and action tokens. The full objective is given in
Appendix~\ref{sec:appendix-grpo}.

Through this trajectory-level optimization, the model learns when to continue searching and when to stop, and it learns to reach a definitive conclusion even under insufficient evidence. The search cost term further discourages redundant retrieval, which reduces the average number of search calls per triple and lowers inference cost at industrial scale.

\section{Experiments}

\begin{table*}[t]
\centering
\small
\setlength{\tabcolsep}{3pt}
\begin{tabular}{lcccccccc}
\toprule
& \multicolumn{4}{c}{\textbf{T-REx long-tail}}
& \multicolumn{4}{c}{\textbf{T-REx unseen}} \\
\cmidrule(lr){2-5} \cmidrule(lr){6-9}
\textbf{Method} & \textbf{P-F1} & \textbf{N-F1} & \textbf{Macro} & \textbf{Calls}
                & \textbf{P-F1} & \textbf{N-F1} & \textbf{Macro} & \textbf{Calls} \\
\midrule
Direct LLM          & 0.598 & 0.714 & 0.656 & 0.00 & 0.571 & 0.745 & 0.658 & 0.00 \\
Single-turn RAG     & 0.747 & 0.806 & 0.777 & 1.00 & 0.759 & 0.822 & 0.791 & 1.00 \\
IRCoT               & 0.819 & 0.848 & 0.834 & 5.39 & 0.813 & 0.842 & 0.828 & 5.58 \\
\midrule
AgentKGV (framework) & 0.822 & 0.842 & 0.832 & 1.82 & 0.838 & 0.855 & 0.847 & 1.86 \\
\quad + SFT          & 0.861 & 0.867 & 0.864 & 3.24 & 0.847 & 0.860 & 0.854 & 3.36 \\
\quad + SFT + GRPO   & \textbf{0.872} & \textbf{0.870} & \textbf{0.871} & 1.63
                     & \textbf{0.864} & \textbf{0.862} & \textbf{0.863} & 1.80 \\
\bottomrule
\end{tabular}
\caption{Main results on the two T-REx splits (long-tail $\rightarrow$
unseen). \textbf{P-F1}/\textbf{N-F1} are per-class F1,
\textbf{Macro} is their macro-average, and \textbf{Calls} is the average search
calls per triple.}
\label{tab:trex_results}
\end{table*}

\subsection{Experimental Settings}

\paragraph{English Open KG}
Our main benchmark is built from T-REx \cite{elsahar2018trex}. The source set contains 50K triples over 376 predicates and we define two disjoint predicate regimes. $P_{\text{long-tail}}$ appears in training but each predicate is capped at 10 triples and $P_{\text{unseen}}$ is held out from training.

\paragraph{Hard negatives.}
We build \emph{hard} negatives rather than random ones. For each positive $(s,p,o)$, we replace $o$ with the same-predicate object $o'$ that is most similar to $o$ under \texttt{BAAI/bge-base-en-v1.5} embeddings \cite{xiao2024cpack}, and we exclude any $(s,p,o')$ that already exists as a positive. A per-split similarity threshold balances the positive and negative counts, and all negatives pass validation without collisions against known positives (Table~\ref{tab:negatives}).

\paragraph{Korean Enterprise KG (industrial application).}
To test real-world transfer, we evaluate on an automatically constructed Korean enterprise KG, where the negatives are genuine real-world errors arising from extraction failures and noise in the source documents. Human annotators labeled each triple, and the final set contains 337 positive and 114 negative triples. Although small in scale, every triple is human-verified, making it a high-quality testbed. This benchmark differs from T-REx in language, domain, and retriever.

\paragraph{Setup.}
The English benchmark uses a BGE dense retriever \cite{xiao2024cpack}, while the enterprise benchmark uses an in-house lexical retriever. We retrieve the top-5 documents per query, cap the number of turns at 8, and summarize with Qwen-2.5-7B-Instruct \cite{qwen2024qwen25}. The search penalty coefficient $\alpha$ is set to 0.05 on T-REx and 0.12 on the Korean enterprise KG. The teacher is gpt-oss-120b, and the backbone is Qwen-2.5-7B-Instruct, shared across all LLM baselines.

\paragraph{Baselines / Metrics.}
We compare against Direct LLM, Single-turn RAG, IRCoT \cite{trivedi2022interleaving}. We report per-class F1 (Pos-F1, Neg-F1), their macro-average (Macro) and the average search calls per triple. For retrieval-based models, if a valid label is not produced before generation terminates, the prediction is scored as negative.

\subsection{Main Results}
\label{sec:main}
Table~\ref{tab:trex_results} shows that our framework outperforms all baselines. Two-stage training achieves the best per-class F1, and it improves over Single-turn RAG by a clear margin on both splits. The gap to IRCoT is also consistent across both splits, even though IRCoT issues substantially more search calls per triple. This pattern indicates that Stage~1 gives a stable query-rewriting foundation that transfers to predicates with limited or no training exposure, rather than to in-distribution relations only.

Two-stage training also cuts retrieval cost. GRPO reduces the average number of search calls consistently across both T-REx splits, while F1 is improved. The reduction is most valuable on the unseen split, where a search agent risks a blow-up on relations it has never encountered, yet our model keeps its search calls controlled.

\subsection{Industrial Applicability}
\label{sec:industrial}
On the Korean enterprise KG (Table~\ref{tab:enterprise_results}), the framework substantially outperforms the baselines, and two-stage training achieves the best Pos-F1 and Macro-F1. This result holds even though the benchmark differs in language, domain, and retriever, and its negatives are real extraction errors. Notably, the untrained framework (AgentKGV) leads on P-F1 and Macro-F1, but its N-F1 still trails IRCoT. Architecture alone does not resolve this noisy negative class. SFT closes this gap and achieves the best N-F1. GRPO then raises P-F1 further, at a small cost to N-F1, and this yields the best overall Macro-F1.

\begin{table}[t]
\centering
\small
\setlength{\tabcolsep}{4pt}
\begin{tabular}{lccc}
\toprule
\textbf{Method} & \textbf{P-F1} & \textbf{N-F1} & \textbf{Macro} \\
\midrule
Direct LLM           & 0.591 & 0.375 & 0.483 \\
Single-turn RAG      & 0.391 & 0.394 & 0.392 \\
IRCoT                & 0.618 & 0.405 & 0.511    \\
\midrule
AgentKGV (framework) & 0.701 & 0.381 & 0.541 \\
\quad + SFT          & 0.745 & \textbf{0.440} & 0.593 \\
\quad + SFT + GRPO   & \textbf{0.794}   & 0.422  & \textbf{0.608} \\
\bottomrule
\end{tabular}
\caption{Results on the Korean enterprise KG (industrial application).}
\label{tab:enterprise_results}
\end{table}

\subsection{Ablation Study}
\label{sec:ablation}
\paragraph{Effect of each training stage.}
Table~\ref{tab:trex_results} disentangles the contribution of each stage. Stage~1 (SFT) consistently improves Macro-F1 over the untrained framework on both splits, confirming that turn-level distillation stabilizes query rewriting and reasoning. However, it does so at the cost of more retrieval: the average search calls rise on every split, because the distilled policy leans on the teacher's retrieve-when-uncertain behavior and searches more aggressively. Stage~2 (GRPO) resolves this trade-off. It cuts the search calls back to near the untrained level while \emph{further} improving Macro-F1, which shows that many of the extra searches induced by Stage~1 were redundant rather than informative.

\section{Conclusion}
We presented an Agentic LLM-RAG framework for KG fact verification, together with a two-stage training strategy that stabilizes query rewriting, optimizes the iterative search policy. Experiments demonstrate that our framework outperforms representative baselines, and the ablation study confirms the complementary contribution of each training stage. We believe this framework offers a practical foundation for reliable fact verification in industrial KGs.

\section{Limitations}
\paragraph{Scale of the industrial benchmark.}
Our industrial evaluation relies on a human-annotated Korean enterprise KG. Because every triple is labeled and verified by human annotators, the annotation cost is high, and the resulting benchmark is necessarily small. We therefore treat it as a high-precision sanity check rather than a large-scale evaluation, and the reported industrial numbers should be interpreted with this limited sample size in mind. Scaling the evaluation to larger industrial KGs without prohibitive annotation cost---for example, through semi-automatic labeling or human-in-the-loop verification---is an important direction for future work.

\bibliography{ref}

\appendix
\section{Component Combinations for Initial Query Writing}
\label{sec:appendix-query-combinations}
Table \ref{tab:query_combinations} shows the example component combinations provided to the agent as a guide. For each turn, the agent selects one combination and writes a single query. The example queries are derived from the triple \textit{(Hamlet, written by, Shakespeare)}, where $p'$ denotes a paraphrased surface form of the predicate.
\begin{table}[h]
\centering
\small
\begin{tabular}{ll}
\toprule
\textbf{Combination} & \textbf{Example Agent Query} \\
\midrule
$s$        & Hamlet \\
$s, p$     & Hamlet written by \\
$s, p, o$  & Hamlet written by Shakespeare \\
$s, p'$    & Hamlet author \\
$s, p', o$ & Hamlet author Shakespeare \\
\bottomrule
\end{tabular}
\caption{Component combinations example provided to the agent as a guide.}
\label{tab:query_combinations}
\end{table}

\section{GRPO Objective}
\label{sec:appendix-grpo}

Within each group of $G$ trajectories, the relative advantage of the $i$-th
trajectory is computed by standardizing its reward against the group statistics
\cite{shao2024deepseekmath}:
\begin{equation}
A_i = \frac{R_i - \mathrm{mean}(\{R_j\}_{j=1}^{G})}
            {\mathrm{std}(\{R_j\}_{j=1}^{G})}.
\end{equation}
The trajectory-level advantage is broadcast to every generated token, and the
policy is updated by maximizing the clipped GRPO objective
\begin{equation}
\begin{aligned}
\mathcal{L}_{\text{GRPO}} =\ &\mathbb{E}\Bigg[
\frac{1}{\sum_i |\tau_i|}
\sum_{i=1}^{G} \sum_{t \in \mathcal{M}_i}
\min\Big( r_{i,t}(\theta)\, A_i,\, \\
&\quad \mathrm{clip}\big(r_{i,t}(\theta),\, 1-\epsilon,\, 1+\epsilon\big)\, A_i
\Big)\Bigg] \\
&- \beta \, \mathbb{E}\Big[ D_{\text{KL}}\big(P_\theta \,\|\, P_{\text{ref}}\big)\Big],
\end{aligned}
\end{equation}
where
$r_{i,t}(\theta) = \dfrac{P_\theta(a_{i,t}\mid s_{i,t})}{P_{\theta_{\text{old}}}(a_{i,t}\mid s_{i,t})}$
is the per-token importance ratio, $\epsilon$ is the clipping range, and
$\mathcal{M}_i$ denotes the set of \emph{model-generated} token positions in
trajectory $\tau_i$. Tokens produced by the retrieval environment are excluded
from $\mathcal{M}_i$.

\section{Hard-Negative Construction Statistics}
\label{sec:appendix-negatives}

\begin{table}[h]
\centering
\small
\setlength{\tabcolsep}{4pt}
\begin{tabular}{lccc}
\toprule
\textbf{Split} & \textbf{Pos / Neg} & \textbf{Thresh.} & \textbf{Mean Sim.} \\
\midrule
test\_long\_tail & 500 / 499   & 0.45 & 0.660 \\
test\_unseen     & 500 / 500   & 0.45 & 0.700 \\
\bottomrule
\end{tabular}
\caption{Hard-negative statistics. \textbf{Mean Sim.}\ is the mean object
cosine similarity between each positive and its negative.}
\label{tab:negatives}
\end{table}

\end{document}